\documentclass[10pt,twocolumn,letterpaper]{article}

\usepackage{float}
\usepackage{booktabs}
\usepackage{colortbl}
\usepackage{tabularx}
\usepackage[linesnumbered,ruled]{algorithm2e}

\usepackage[pagenumbers]{iccv} 

\definecolor{iccvblue}{rgb}{0.21,0.49,0.74}
\usepackage[pagebackref,breaklinks,colorlinks,allcolors=iccvblue]{hyperref}
\usepackage[accsupp]{axessibility}
\def\paperID{4070} 
\def\confName{ICCV}
\def\confYear{2025}

\title{LINR-PCGC: Lossless Implicit Neural Representations for Point Cloud Geometry Compression}

\author{Wenjie Huang$^{1}$ \quad Qi Yang$^{2}$ \quad Shuting Xia$^{1}$ \quad He Huang$^{1}$ \quad Zhu Li$^{2}$ \quad Yiling Xu$^{1\dagger}$ \\
$^{1}$ Shanghai Jiao Tong University\quad \textsuperscript{2} University of Missouri-Kansas City  \\
{\tt\small $^{1}$\{huangwenjie2023, xiashuting, huanghe0429, yl.xu\}@sjtu.edu.cn , $^{2}$\{qiyang, lizhu\}@umkc.edu}}

\begin{document}
\maketitle
\begin{abstract}
Existing AI-based point cloud compression methods struggle with dependence on specific training data distributions, which limits their real-world deployment. Implicit Neural Representation (INR) methods solve the above problem by encoding overfitted network parameters to the bitstream, resulting in more distribution-agnostic results. However, due to the limitation of encoding time and decoder size, current INR based methods only consider lossy geometry compression. In this paper, we propose the first INR based lossless point cloud geometry compression method called Lossless Implicit Neural Representations for Point Cloud Geometry Compression (\textbf{LINR-PCGC}). To accelerate encoding speed, we design a group of point clouds level coding framework with an effective network initialization strategy, which can reduce around 60\% encoding time. A lightweight coding network based on multiscale SparseConv, consisting of scale context extraction, child node prediction, and model compression modules, is proposed to realize fast inference and compact decoder size. Experimental results show that our method consistently outperforms traditional and AI-based methods: for example, with the convergence time in the MVUB dataset, our method reduces the bitstream by approximately 21.21\% compared to G-PCC TMC13v23 and 21.95\% compared to SparsePCGC. Our project can be seen on \href{https://huangwenjie2023.github.io/LINR-PCGC/}{https://huangwenjie2023.github.io/LINR-PCGC/}.

\end{abstract}    
\begin{figure}[t]
  \centering
  \includegraphics[width=\linewidth]{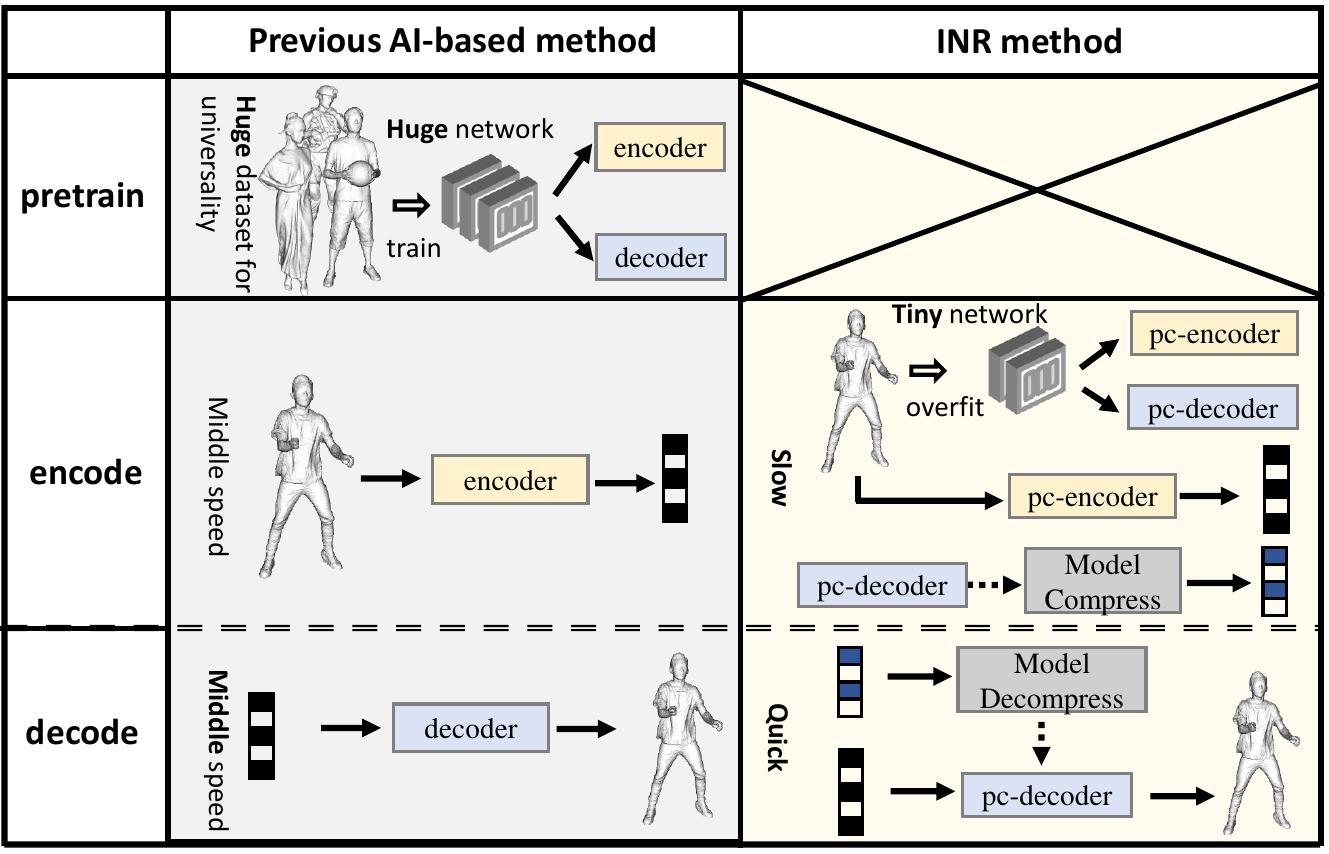}
   \caption{Previous AI-based method typically employs large-scale datasets to train a high-capacity neural network for universality, which operates in inference mode with fixed parameters during both encoding and decoding. INR method overfits the network with target point cloud to be compressed. Both the point cloud and the network parameters will be encoded into bitstream.}
   \label{fig:compare}
\end{figure}

\section{Introduction}
\label{sec:intro}

Point clouds have become a pivotal data format for representing and interacting with 3D environments. Their ability to capture complex spatial structures with high fidelity makes them indispensable for many real-world applications, such as applications in the metaverse and autonomous driving~\cite{pc_survey_cao, quach2022survey}. However, the large data volume of point clouds, specifically for the point cloud sequence that can capture dynamic content,  poses significant challenges for storage, transmission, and real-time processing, necessitating the development of efficient compression techniques. In this paper, we focus on the Point Cloud sequence Geometry Lossless Compression (PCGLC) study.

Current PCGLC methods can be mainly categorized into traditional approaches and AI-driven approaches. Traditional methods, such as Geometry-based Point Cloud Compression (G-PCC)~\cite{GPCC_method, gpcc_codec} and Video-based Point Cloud Compression (V-PCC)~\cite{V-PCC, vpcc_codec, emerging}, have been standardized by the Coding of 3D Graphics and Haptics (WG7) of the Moving Picture Experts Group (MPEG). These methods have demonstrated strong adaptability, interpretable operations, and impressive compression ratios. Despite these advantages, these methods rely on manually designed tools and parameters, which limit their ability to fully exploit geometry spatial, leading to suboptimal performance~\cite{pc_survey_cao}. AI-driven methods, based on 3D regular voxels~\cite{nguyen2021multiscale, unicorn_geo, fan2023multiscale} or tree structures~\cite{huang2020octsqueeze, muscle, octattention, song2023efficient}, address these limitations using neural networks to model spatial correlation from high-dimensional latent spaces. These approaches achieve state-of-the-art (SOTA) performance on specific datasets. However, they heavily rely on the training datasets, sample distribution shifts would result in significant performance drops, consequently limiting their practical application. 

To achieve more stable compression performance for AI-driven methods, Implicit Neural Representation (INR) based compression methods~\cite{wangyao_nvfpcc, isik2022lvac, SCNIR, lightweight_mlp, pcc_implicit} are proposed to generate overfitted encoder and decoder for target samples. Compared to previous AI-based methods, the INR method offers significant advantages in adaptability. In addition, they do not require a large number of training samples with various attributes, as shown in \cref{fig:compare}. However, the INR method has two main challenges: 1) the network parameters, especially for the decoder, need to be encoded into bitstreams. Therefore, current INR methods prefer to use simple networks, which generally have a relatively weak fitting capability and are therefore limited to lossy compression~\cite{isik2022lvac, SCNIR}, resulting in the emptiness of INR based PCGLC, and 2) the overfitting time for the network, which is considered part of the encoding time of INR methods, is too long to limit the potential for real applications~\cite{wangyao_nvfpcc, pcc_implicit, lightweight_mlp}.
    


To solve the above problems and fill the gap in PCGLC in terms of INR, we propose a novel framework called Lossless Implicit Neural Representation for Point Cloud Geometry Compression (\textbf{LINR-PCGC}). For the first problem, we draw inspiration from the Group of Pictures (GoP) concept in video encoding: the adjacent frames from a point cloud sequence should have close characteristics, and these frames can share a common lightweight network for encoding and decoding\footnote{Adjacent frames are only used to share the network parameters. We have not used inter prediction, the order of frames has no impact on our method.}. By doing so, we can reduce the average bandwidth cost of the network parameters. The GoP-wise operation also leads to an initialization strategy for solving the second problem: the network trained in the previous GoP can be used as the initialized network for overfitting of the next GoP.

The lightweight network used for encoding and decoding is based on multiscale SparseConv \cite{minkowskiengine}: continuously downsampling until there are only a few dozen or a few hundred points (high scale to low scale), followed by an effective upsampling network called Child Node Prediction (CNP) to estimate the occupancy probability of the higher scale point. Because all scales of a point cloud share the same set of network parameters, we propose the Scale Context Extraction (SCE) module to distinguish different scales and improve compression efficiency. To generate a compact network bitstream, we propose an Adaptive Quantization (AQ) module and a Model Compression (MC) module to quantize and encode the network parameters. A regularization term is introduced for the optimization process to make the training process more stable and the network parameters easier to compress. Our main contributions are as follows. 
\begin{itemize}
    \item We propose the first INR based method for PCGLC called LINR-PCGC. A lightweight multiscale SparseConv network is designed to realize effective point cloud lossless compression.
    \item Our method allows a group of frames to share a common decoder, reducing the bandwidth cost of the network. And we design an initialization strategy based on GoP which has been shown to save about 65.3\% encoding time.
    \item The results of the experiments show that the proposed method reports superior performance to traditional and AI-driven SOTA methods.
\end{itemize}

\section{Related Works}
\subsection{Traditional methods} Two main traditional point cloud compression methods are Geometry-based Point Cloud Compression ({G-PCC}) and Video-based Point Cloud Compression ({V-PCC}). G-PCC directly encodes point clouds in 3D space using hierarchical structures like octrees, which recursively subdivide the 3D volume to represent point locations. This method is particularly effective for sparse point clouds, such as LiDAR data~\cite{pc_survey_Graziosi}. However, G-PCC can be computationally demanding due to the need for complex 3D data processing and the lack of efficient temporal prediction for dynamic sequences. On the other hand, V-PCC transforms 3D point clouds into 2D images by segmenting the point cloud into patches, projecting them onto 2D planes, and packing these patches into images for compression using existing video codecs like HEVC. This approach leverages the efficiency and real-time decoding of 2D video compression and is well-suited for dense, dynamic point clouds~\cite{pc_survey_cao, vpcc_paper}, but struggles with sparse point clouds.

\subsection{AI-based methods} PCGCv2~\cite{PCGCV2} employs a multiscale end-to-end learning framework that hierarchically reconstructs point cloud geometry through progressive re-sampling. It uses sparse convolution neural network (SparseCNN) based autoencoders~\cite{balle2016end, balle2018variational} to compress binary occupancy attributes into downscaled point clouds with geometry and feature attributes. The lowest scale geometry is losslessly compressed using an octree codec, while feature attributes are lossy compressed using a learned probabilistic context model. Subsequently, SparsePCGC~\cite{sparsepcgc} introduces a unified framework based on multiscale sparse tensor representation. It processes only the most probable positively occupied voxels (MP-POV) using sparse convolutions, reducing computational complexity. SparsePCGC incorporates a SparseCNN-based Occupancy Probability Approximation (SOPA) model to estimate occupancy probabilities by exploiting cross-scale and same-scale correlations. Additionally, it uses SparseCNN-based Local Neighborhood Embedding (SLNE) to aggregate local variations as spatial priors, further enhancing compression efficiency. SparsePCGC achieves excellent performance in both lossless and lossy compression across several datasets ruled by MPEG, including dense objects and sparse LiDAR point clouds. These methods demonstrate the potential of AI-driven techniques in addressing the challenges of point cloud compression. And the state-of-the-art AI-based geometry point cloud compression method {Unicorn-Part I}~\cite{unicorn_geo} is based on SparsePCGC. There are two reasons why we use SparsePCGC instead of Unicorn-Part I as the baseline: 1) Unicorn-Part I contains many parameters and a more complex network structure, making model simplification very difficult, which is not conducive to initial exploration due to its complexity. 2) Unicorn-Part I is based on SparsePCGC, so subsequent improvements are compatible with our framework construction based on SparsePCGC.


\section{Method}
\begin{figure*}[t]
  \centering
  \includegraphics[width=\textwidth]{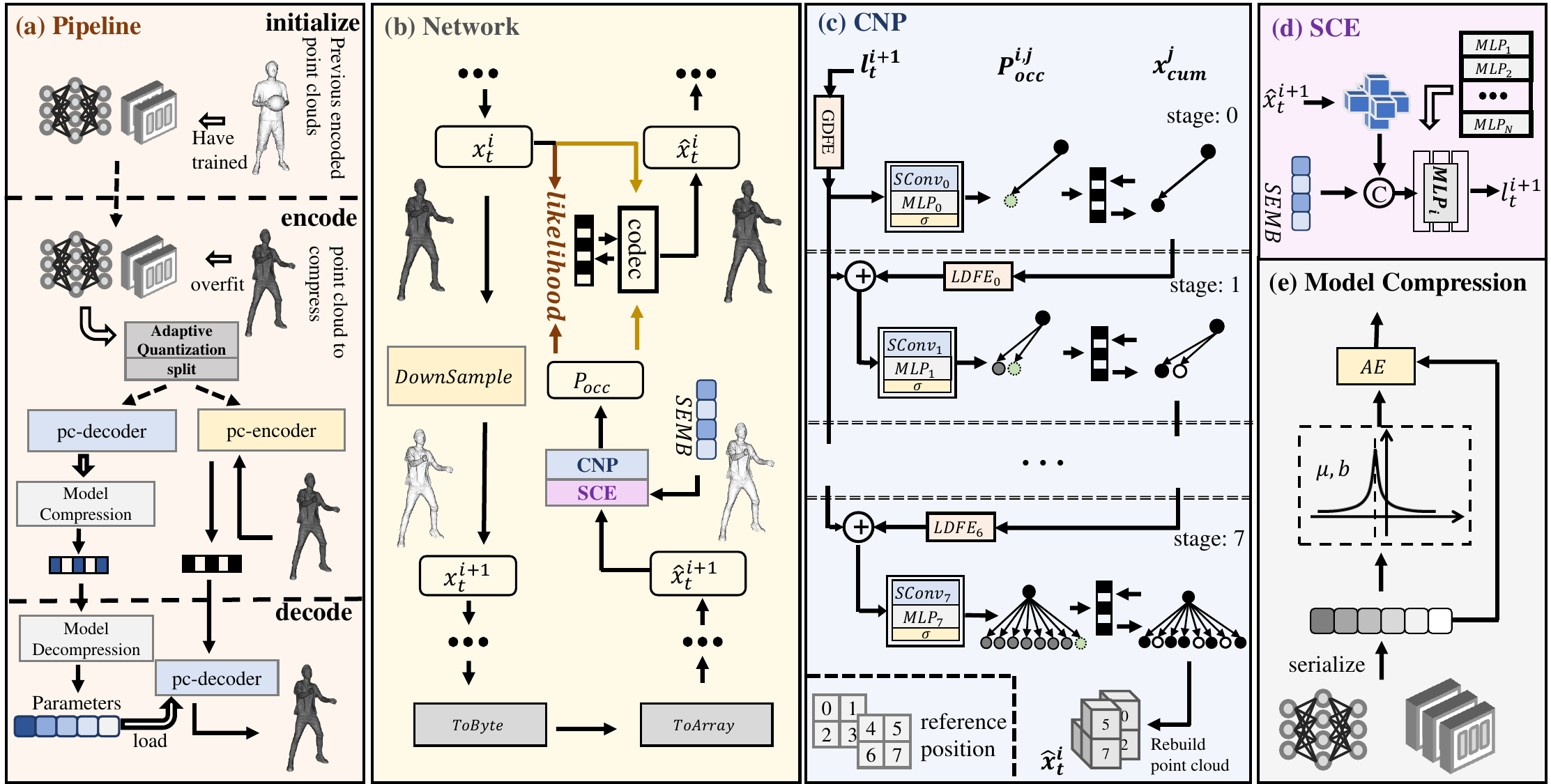}
    \caption{LINR-PCGC Framework. (a) Pipeline. (b) Network. (c) Child Node Prediction. (d) Scale Context Extraction.  (e) Model Compression.}
  \label{fig:overall}
\end{figure*}

\subsection{Pipeline}
As shown in {\cref{fig:overall} (a)}, we design a multiscale network for point cloud encoding and decoding. 3 steps are required to encode a GoP: 
\begin{itemize}
    \item Initialize. Initialize network by network has been overfitted by the previous GoP. The first GoP is initialized randomly.
    \item Encode. Overfit the initialized network with target point clouds that need to be compressed. Next, split the model into the point cloud encoder (pc-encoder) and point cloud decoder (pc-decoder). Finally, encode the point cloud with the pc-encoder and encode the pc-decoder parameters with the AQ and MC module.
    \item Decode. Decode the model parameters of the pc-decoder from the Model Decompress (MD) module. Then, decode point clouds using the pc-decoder.
\end{itemize}
 
\subsection{Initialization Strategy}
The encoding process is carried out on a GoP-wise basis, treating each GoP as a unit of compression. Then, there can be a useful initialization strategy: initialize the network of the current GoP with the network overfitted by the previous GoP. This approach significantly accelerates the overfitting process. For the first GoP, we can either initialize it randomly or using other similar content, which will be discussed in \cref{sec:ab_is}.


\subsection{Network}

Let $S = \{x_1, \dots, x_t,\dots, x_M\}$ represent a point cloud sequence with $M$ frames, where $x_t = \{C_t, F_t\}$ represents a single point cloud frame with a time index $t$, $C_t$ represents the coordinates of occupied points in $x_t$, and $F_t$ represents its associated attributes\footnote{Since this work focuses on the compression of point cloud coordinates, $F_t$ of the initialized point cloud is a vector of ones.}. The point cloud sequence $S$ can be uniformly grouped into multiple GoPs: $G_1 = \{x_1, \dots, x_T\}$, $G_2 = \{x_{T+1}, \dots, x_{2T}\}$, $\dots$, $G_r = \{x_{(r-1)T+1}, \dots, x_M\}$.

The basic network architecture is shown in \cref{fig:overall} (b). First, progressively downsample the point cloud until there are only a few dozen or a few hundred points in the lowest scale point cloud. Then, the spatial position of the low-scale point cloud is used to predict the occupancy information of the high-scale point cloud. Each scale has a bitstream to encode the occupancy information using the predicted occupancy probability. Next, the lowest point cloud is converted to bytes directly because there are very few points left, and the decoder network is compressed by the MC module. Finally, the lowest scale point cloud information, the decoder network parameter information, and the occupancy information of each scale make up the final bitstream.

\subsubsection{Point Cloud Downsampling}
Downsampling is used to express the approximate structure of a point cloud using a lower resolution point cloud. 
\begin{equation}
    x_t^{i+1} = DS(x_t^{i})
\end{equation}
Here, we use $DS (\cdot) = Maxpooling (\cdot)$ to do the downsampling. 

\subsubsection{Scale Context Extraction}
To distinguish the different spatial scales of point clouds, we design a module called SCE, depicted in \cref{fig:overall} (d). SCE considers scale embedding (SEMB) as global information and neighbor occupancy as local information. For each scale, an MLP is used after concatenating global information and local information to generate scale context features, which can be formulated as:
\begin{align}
    Nb^{i+1} &= \mathcal{F
    }(\hat{x}_t^{i+1}) \\
    l_t^{i+1} &= MLP_i(Concate(Nb^{i+1}, SEMB(i)))
\end{align}
where $\mathcal{F}$ denotes finding the occupancy of the ``front, behind, left, right, up, down, self" position for each point. $SEMB(i)$ denotes a scale embedding that uses an 8-channel implicit feature to extend the scale index $i$. $Concate(Nb^{i+1}, SEMB)$ denotes channel connection for $Nb^{i+1}$ and $SEMB$. $MLP_i$ denotes multilayer perceptron for merging global and local information of the $(i+1)$th scale to derive $l_t^{i+1}$, which denotes the scale information of the current scale and will be used in the following introduced {CNP} module.

\subsubsection{Child Node Prediction}
\begin{figure}[htp]
    \centering
    \includegraphics[width=\linewidth]{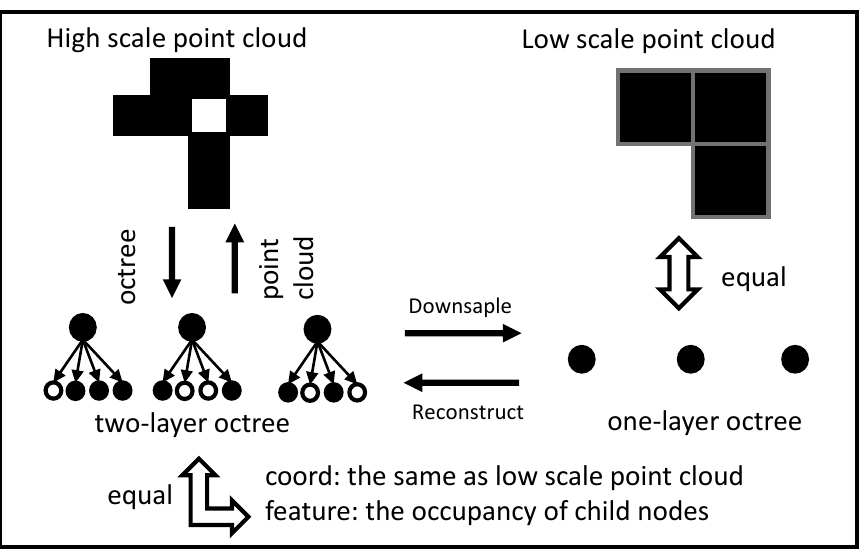}
    \caption{Toy example of CNP.}
    \label{fig:octree}
\end{figure}

Child Node Prediction (CNP) is designed to upsample point clouds from a lower to a higher spatial scale. The previous method~\cite{sparsepcgc} uses transpose convolution to upsample the point cloud, which incurs high memory usage and time complexity. So we propose a new method for upsampling: seeking the child node of the octree as depicted in \cref{fig:octree}.

A high-scale point cloud can be used to establish a two-layer octree. The two-layer octree equals a point cloud in which the coordinates are the same as the low-scale point cloud, and the features are the occupancy of the child nodes. In \cref{fig:octree}, we replace the octree in a 3D space with a quadtree in a 2D plane for simplicity, so the occupancy channel in \cref{fig:octree} has a value of 4, while in the octree, the occupancy channel is 8. The downsampling can be seen as removing the feature of the two-layer octree. The one-layer octree and the low-scale point cloud are the same. Reconstruction of the high-scale point cloud equals reconstructing the \textbf{occupancy of the child nodes}.  

\begin{figure}[ht]
    \centering
    \includegraphics[width=\linewidth]{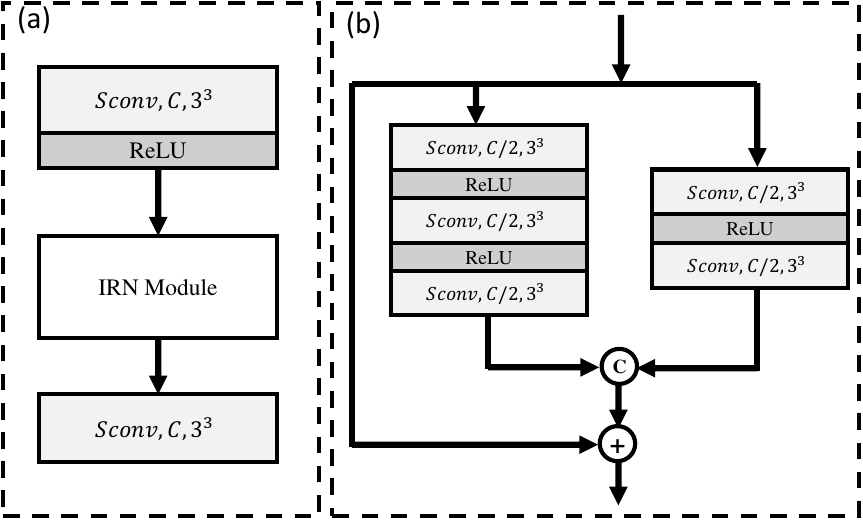}
    \caption{(a) The structure of GDFE/LDFE. (b) The structure of IRN Module. $Sconv, C, k^3$ denotes a Sparse convolution layer with C output channels and kernel size $k$, respectively.}
    \label{fig:DFE}
\end{figure}


To improve the accuracy of the prediction and reduce the bitstream, we predict the occupancy of the child nodes in a channel-wise rule. The decoded child nodes serve as context information for the child nodes to be decoded. We use an 8-stage method to reconstruct a high-scale point cloud as shown in \cref{fig:overall} (c).  Let $j$ ($0,1,2,...,7$) represent the stage index. $x_{cum}^j$ represents the reconstructed point cloud in $j$th stage. $P_{occ}^{i,j}$ represents the predicted probability of each child node in $i$th scale $j$th stage. $l_t^{i+1}$ denotes the $(i+1)$th scale point cloud with scale information derived from the SCE module. Two feature extraction modules, GDFE (Global Deep Feature Extraction) and LDFE (Local Deep Feature Extraction), are proposed as shown in \cref{fig:DFE}.


We first use GDFE and LDFE modules to extract two latent features, i.e., $G_{feat}$ and $L_{feat}^{j-1}$. Then we merge two features as input to a classification network, consisting of a sparse convolution $Sconv_j$, a multilayer perceptron $MLP_j$ and a nonlinear activation unit $\sigma$ ($Sigmoid$), to obtain $P_{occ}^{i,j}$ that describes the probability of child node occupancy. Specifically, $Sconv_j$ is to extract local neighborhood information, $MLP_j$ is to efficiently integrate point-wise features, and $Sigmoid$ is used to normalize the features to ensure that the occupancy probability falls within the range [0,1]. Finally, the estimated occupancy probability $P_{occ}^{i,j}$ is used for arithmetic coding of the ground truth occupancy values $x^{i,j}_t$. Meanwhile, the cross entropy between $x^{i,j}$ and $P_{occ}^{i,j}$ is calculated to estimate the bitstream size for encoding $x^{i,j}$ with $P_{occ}^{i,j}$, where $x^{i,j}_t$ is the ground truth in $i$-th scale $j$-th stage. The detailed process of CNP during training can be found in \cref{alg:cnp}.

\begin{algorithm}
\caption{Process of CNP during training}
\label{alg:cnp}
\KwIn{$l_{t}^{i+1}$ extracted from SCE.}
\KwOut{Estimated bitstream size $E$ of point cloud.}

Calculating the global feature of all stages, $G_{\text{feat}}=GDFE(l_{t}^{i+1})$;

let bitstream size $E=0$;

\For{each $j$ in $0,\ldots,7$}{
    \If {$j==0$}{
        Use global feature to represent the merge feature directly, $F_{\text{merge}}^{j}=G_{\text{feat}}$;
    }
    \Else{
        Derive local feature from decoded child node occupancy by LDEF, $L_{\text{feat}}^{j-1}=LDEF_{j-1}(x_{\text{cum}}^{j-1})$; 
        
        Add global feature and local feature together to get the merge feature, $F_{\text{merge}}^{j}=G_{\text{feat}}+L_{\text{feat}}^{j-1}$;
    }

    Calculate the child node occupancy probability $P_{occ}^{i, j} = \sigma(MLP_{j}(SConv_{j}(F_{merge}^{j})))$;
    
    Use entropy to estimate the bitstream size of current stage, $E_{j}=Entropy(x_{t}^{i, j}, P_{occ}^{i, j})$;
    
    Cumulative the bitstream size, $E=E+E_{j}$;
    
    \If{$j==0$}{
        Record the decoded child node occupancy $x_{cum}^{j}=x_{t}^{i, j}$;
    }
    \Else{
        Accumulated the decoded child node occupancy $x_{cum}^{j}=Concate(x_{cum}^{j-1}, x_{t}^{i, j})$, $Concate$ represents channel connection;
    }

}

\end{algorithm}



\subsection{Adaptive Quantization}
To make it easier to encode the network parameters, we need to quantize them using the AQ module. Let $p_{dec}$ represent the parameters related to the pc-decoder.
\begin{align}
normalize(p_{dec}) = \frac{p_{dec} - \min(p_{dec})}{\max(p_{dec}) - \min(p_{dec})} \\
p_{quant} = round(normal(p_{dec}) \times (2^B - 1))
\end{align}
Where $normal$ is a normalization for $p_{dec}$ to ensure its values within the range $[0,1]$. Then, it multiplies by $2^B-1$, followed by a round operation, to quantize it to $B$ bits. The dequantization is:

\begin{equation}
    p_{dequant} = \frac{p_{quant}}{2^B-1}\times(\max{(p_{dec})}-\min{(p_{dec})}) + \min{(p_{dec})}
\end{equation}


\subsection{Model Compression}
\begin{figure}[ht]
    \centering
    \includegraphics[width=\linewidth]{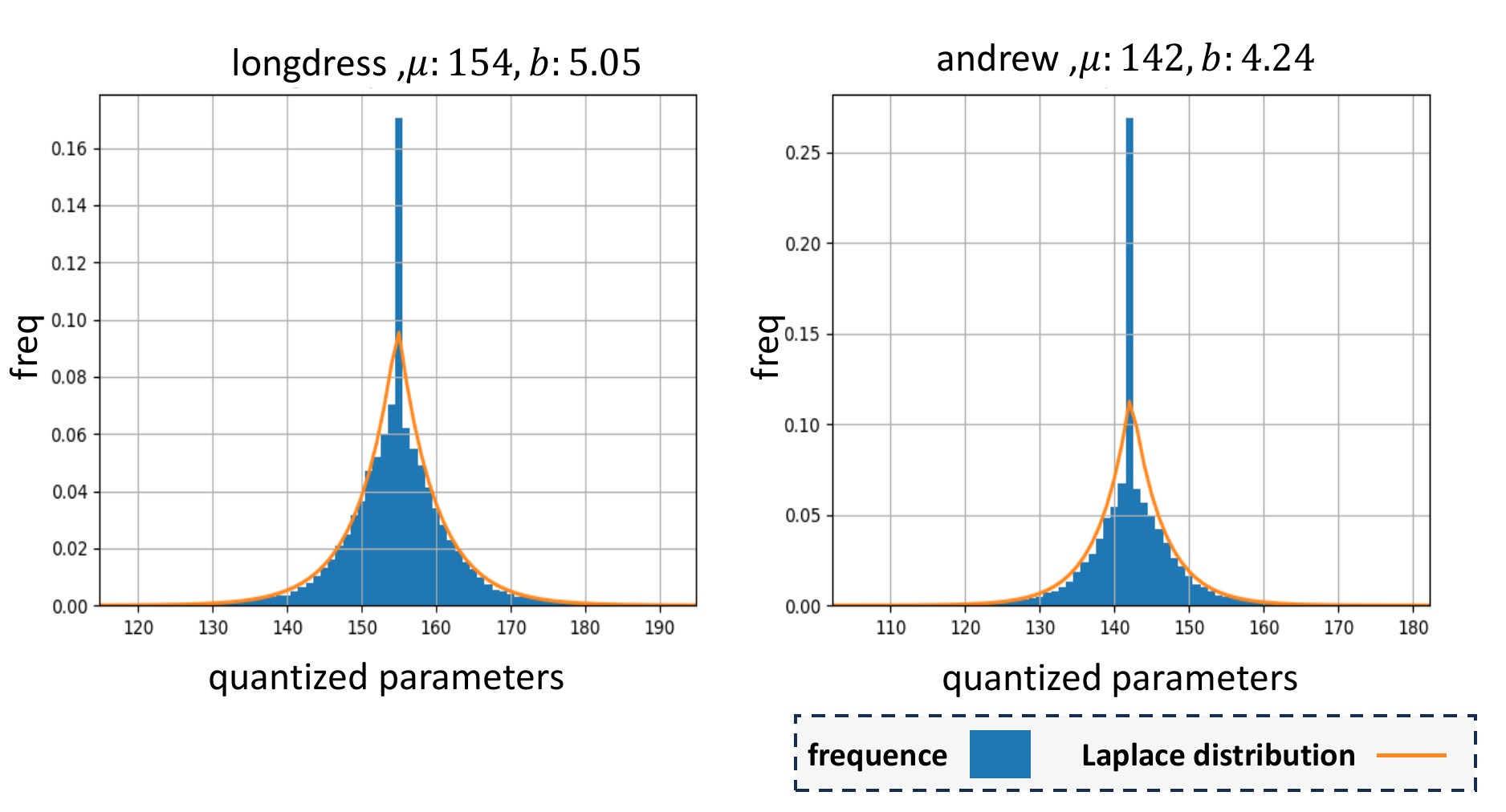}
    \caption{Distribution of quantized network parameters.}
    \label{fig:distribution}
\end{figure}

To better understand the parameter distribution of the model, we plot and analyze its frequency histogram in \cref{fig:distribution}. It is evident that the model parameters trained with regularization terms closely follow a Laplace distribution. Consequently, in quantized model parameters' entropy coding, we employ a Laplace distribution as an approximation of their actual distribution, denoted as $f$ in \cref{eq:laplace3}. The mean and scale parameters are estimated using \cref{eq:laplace} and \cref{eq:laplace2}, which are encoded into bitstream as side information.
\begin{align} 
    &f(p_{quant}; \mu, b) = \frac{1}{2b} \exp \left( -\frac{|p_{quant} - \mu|}{b} \right) \label{eq:laplace3} \\
        &\mu = \frac{1}{n}\sum(p_{quant}) \label{eq:laplace} \\
    &b = \frac{1}{n}\sum{|p_{quant}-\mu|} \label{eq:laplace2}
\end{align}

\subsection{Loss Function}
The loss of the training process is derived from the CNP module, which calculates the cross entropy between $x^{i,j}_t$ and $P_{occ}^{i,j}$ of each stage on each spatial scale as an estimation of bitstream size. As the ground truth occupancy $x^{i,j}_t$ is binary, the Binary Cross-Entropy (BCE) is used,
\begin{align}
    L_{BCE}^{i,j} &= Entropy(x^{i,j}_t,P_{occ}^{i,j}) \\
    \mathcal{L} &= \sum_{i=0}^{N}\sum_{j=0}^{7}L_{BCE}^{i,j} + \lambda||\boldsymbol{\theta}||^2_2
\end{align}
where $Entropy(p,q) = -(p\log_2(q)+(1-p)\log_2(1-q))$, $\boldsymbol{\theta}$ represents network parameters, $\lambda$ is an L2 regularization coefficient. Since we only use multi-frames to share network parameters without utilizing inter-frame features, each frame in a GoP can be optimized separately, and there is no need to sum in the $t$ dimension.

\section{Experiment}
\subsection{Experiment Configuration}
\textbf{Training.} Our model is implemented in Pytorch~\cite{paszke2019pytorch} and MinkowskiEngine~\cite{minkowskiengine}. The number of spatial scales is determined by the downsampling times of the first frame until its point number is less than or equal to 64. We use the Adam optimizer~\cite{kingma2015adam} with a learning rate decayed from 0.01 to 0.0004. We set the number of frames per sequence to 96, the GoP size to 32 and the bit depth $B$ in AQ to 8. Further details on hyperparameters are provided in the Appendix. We train the first GoP for 6 epochs and subsequent GoPs for 1 to 6 epochs on a single NVIDIA RTX 3090 GPU with AMD EPYC 7502 CPUs.

\textbf{Dataset.} 
\begin{figure}
    \centering
    \includegraphics[width=\linewidth]{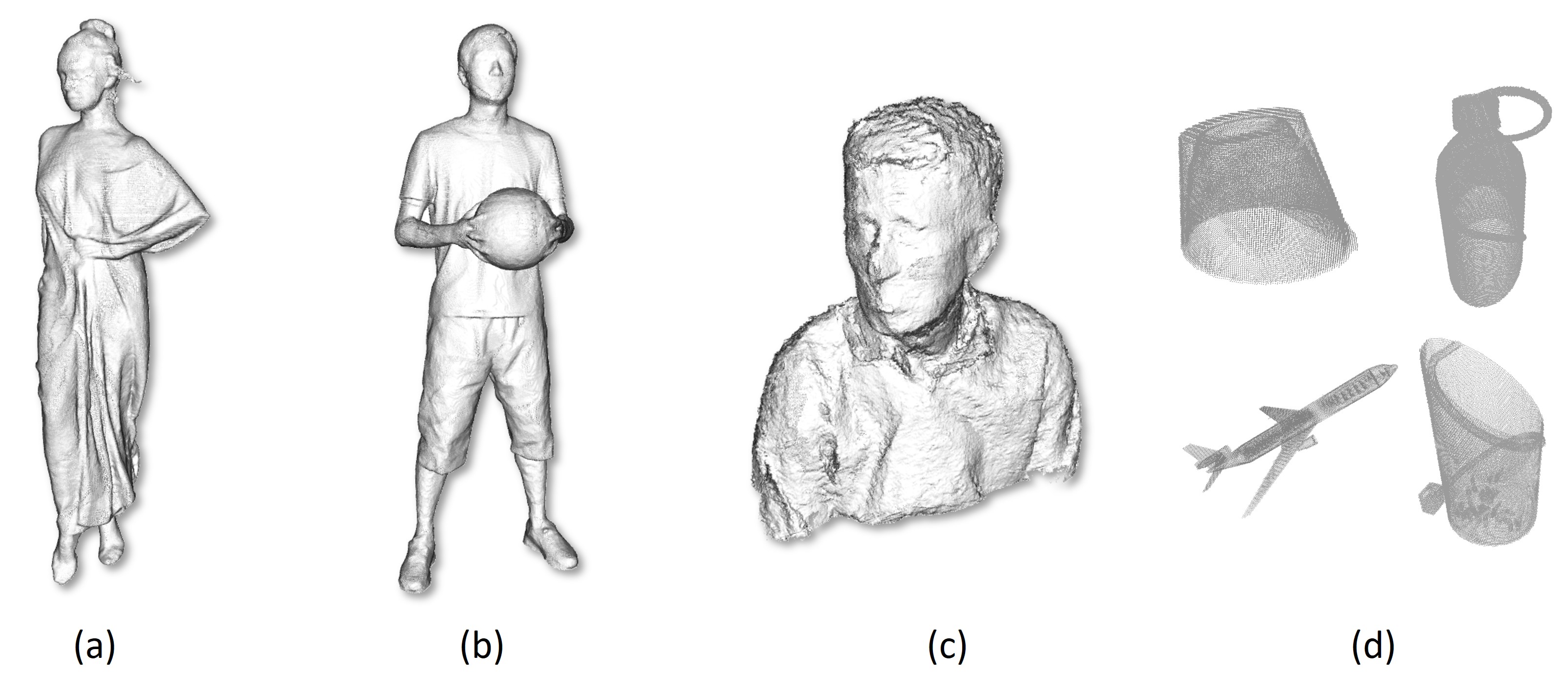}
    \caption{(a) Longdress from 8iVFB. (b) Basketball Player from Owlii. (c) Andrew from MVUB. (d) Samples in Shapenet.}
    \label{fig:dataset}
\end{figure}
As seen in \cref{fig:dataset}, we utilize three different dynamic human datasets with geometric bit-depth 10: 1) 8i Voxelized Full Bodies (8iVFB) dataset~\cite{8iVFB}, which contains 4 sequences at a frame rate of 30 fps over 10 seconds; 2) Owlii Dynamic Human DPC (Owlii)~\cite{Owlii}, which has 4 sequences with 30 fps over 20 seconds; 3) Microsoft voxelized upper bodies (MVUB)~\cite{MVUB}, which provides 5 sequences of upper body movements. We select the first 96 frames from each sequence in the above datasets.

\textbf{Baseline.} For traditional compression methods, we select G-PCC v23~\cite{gpcc_software} and V-PCC v23~\cite{vpcc_software} as baselines. For AI-based methods, we choose SparsePCGC. For fairness, we use pre-trained models trained on ShapeNet provided by the authors~\cite{sparsepcgc_online}.

\textbf{Metric.} For the lossless compression of point cloud geometric information, our primary metric is the average bitstream size per point, denoted as bits per point (bpp). In addition, we evaluate the encoding and decoding times to assess the computational efficiency of the compression process. Furthermore, we present how the bitstream size varies with encoding time, providing insights into the trade-off between compression efficiency and computational resources. And we randomly initialize the first GoP of each sequence.


\subsection{Experiment Result} \label{sec:exp_result}
\begin{figure*}[t]
    \centering
    \includegraphics[width=\linewidth]{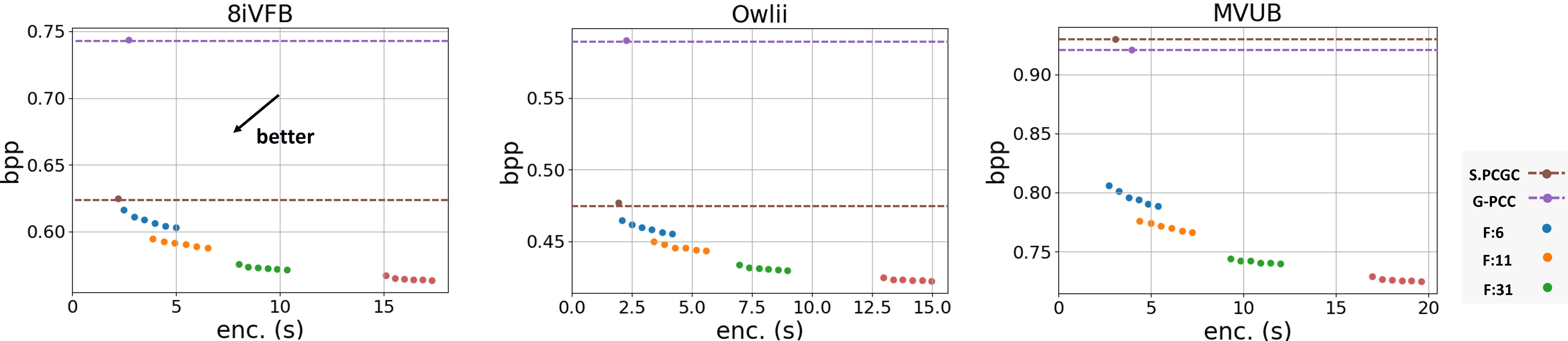}
    \caption{Encoding times vs. bpp curves with different training epochs for the first GoP and subsequent GoPs.}
    \label{fig:bpp_time}
\end{figure*}

\cref{fig:bpp_time} shows the encoding times vs. bpp curves for the 8iVFB, Owlii, and MVUB datasets. As the first GoP is randomly initialized, we tend to train more epochs, i.e., $F$,  for the first GoP. We choose different $F$ values in \cref{fig:bpp_time}, and the different points with the same $F$ denote 1 to 6 training epochs for subsequent GoPs. We can observe that a longer encoding time can achieve a higher compression ratio. In order to observe the compression effect under a comparable encoding time state and the compression effect under a relatively sufficient encoding time state, we collect the first and last points in the time dimension of each figure as quantitative results and construct \cref{tab:comp_8i,tab:comp_owlii,tab:comp_mvub}.
The columns in tables called ``ours" and ``ours 2" are the first and last sampling points (e.g., training 1 and 6 epochs for subsequent GoPs), respectively.

According to the quantitative results, we can observe that: 1) our method can achieve the best compression ratio in a comparable encoding time for all datasets; 2) our method can maintain a fast decoding time, which can be about half that of G-PCC or SparsePCGC (marked as S.PCGC); 3) our method can maintain stable compression performance on datasets with various coordinate distributions. Especially, in MVUB our method achieves a 13.33\% gain compared to SparsePCGC. \cref{fig:dataset} shows coordinate distributions of different datasets, where MVUB has fewer smooth surfaces and more virtual edges than other datasets. Existing AI-based methods like SparsePCGC exhibit poor adaptability in handling large data distribution shifts. 4) If there is enough encoding time (about ten to twenty seconds to encode a frame on average), a larger compression ratio can be obtained, which reduces the bitstream by approximately 9.70\% in 8iVFB, 11.40\% in Owlii, 21.95\% in MVUB than SparsePCGC; about 24.11\%, 28.40\%, 21.21\% than G-PCC.

\begin{table}[tbp]
\centering
\small
\begin{tabular}{lccccc}
\toprule
 & \textbf{G-PCC} & \textbf{S.PCGC} & \textbf{V-PCC} & \textbf{ours} & \textbf{ours 2} \\
\midrule
longdress & 0.74 & 0.619 & 1.384 & 0.618& 0.571
\\
loot & 0.69 & 0.586 & 1.27 & 0.57& 0.521
\\
red\&black & 0.81 & 0.676 & 1.539 & 0.689& 0.626
\\
soldier & 0.734 & 0.619 & 1.469 & 0.588& 0.538
\\
\midrule
bpp (avg)& 0.743 & 0.625 & 1.415 & 0.616 & 0.564
\\
r.t. bpp & 100& 84.044& 190.411
& \cellcolor[HTML]{B2B2FF}82.925& \cellcolor[HTML]{E9AFAA}75.894\\
\midrule
\hline
w/o over.& -& -& -& 0.477& 0.434\\
enc. time & 2.72 & \cellcolor[HTML]{E9AFAA}2.202 & 194.261 & \cellcolor[HTML]{B2B2FF}2.464& 16.423\\
dec. time & 0.923 & 1.048 & 2.304 & \cellcolor[HTML]{B2B2FF}0.501 & \cellcolor[HTML]{E9AFAA} 0.459\\
\bottomrule
\end{tabular}
\caption{Quantitative results on 8iVFB dataset, where bpp (avg) denotes the average bpp of all sequences in the dataset, r.t. bpp denotes the relative bpp (\%) of other methods over G-PCC, and w/o over. denotes the encoding time without overfitting time of our method. All the times are in seconds. The best and the second best results are denoted by red and blue.}
\label{tab:comp_8i}
\end{table}


\begin{table}[tbp]
\centering
\small
\begin{tabular}{lccccc}
\toprule
 & \textbf{G-PCC} & \textbf{S.PCGC} & \textbf{V-PCC} & \textbf{ours} & \textbf{ours 2}\\
\midrule
basketball & 0.578 & 0.466 & 1.097 & 0.452&  0.411
\\
dancer & 0.606 & 0.485 & 1.192 & 0.473&  0.431
\\
exercise & 0.585 & 0.472 & 1.104 & 0.460&  0.417
\\
model & 0.592 & 0.485 & 1.155 & 0.475&  0.431
\\
\midrule
bpp (avg)& 0.59 & 0.477 & 1.137 & 0.465&  0.423
\\
r.t. bpp & 100& 80.815& 192.683
& \cellcolor[HTML]{B2B2FF} 78.759& \cellcolor[HTML]{E9AFAA}71.599\\
\midrule
\hline
w/o over.&  -&    -   &   -      &   0.402&  0.395\\
enc. time   & 2.24   & \cellcolor[HTML]{E9AFAA} 1.932& 146.295   & \cellcolor[HTML]{B2B2FF} 2.493&  14.174\\
dec. time   & 0.725  & 0.926  & 1.985     & \cellcolor[HTML]{B2B2FF} 0.422& \cellcolor[HTML]{E9AFAA} 0.415\\
\bottomrule
\end{tabular}
\caption{Quantitative results on Owlii dataset. The best and the second best results are denoted by red and blue.}
\label{tab:comp_owlii}
\end{table}


\begin{table}[htbp]
\centering
\small
\begin{tabular}{lccccc}
\toprule
 & \textbf{G-PCC} & \textbf{S.PCGC} & \textbf{V-PCC} & \textbf{ours} & \textbf{ours 2} \\
\midrule
andrew10 & 0.941 & 0.947 & 1.611 & 0.833 & 0.748
\\
david10 & 0.898 & 0.909 & 1.462 & 0.778 & 0.704
\\
phil10 & 0.969 & 0.966 & 1.636 & 0.841 & 0.768
\\
ricardo10 & 0.904 & 0.925 & 1.519 & 0.802 & 0.705
\\
sarah10 & 0.892 & 0.901 & 1.486 & 0.777 & 0.702
\\
\midrule
bpp & 0.921 & 0.93 & 1.543 & 0.806 & 0.725
\\
r.t. bpp& 100& 100.947& 167.561
& \cellcolor[HTML]{B2B2FF} 87.548& \cellcolor[HTML]{E9AFAA}78.788\\
\midrule
\hline
w/o over. & -&    -     &     -    &    0.524 & 0.513\\
enc. time & 3.951 & \cellcolor[HTML]{B2B2FF} 3.06& 213.192 & \cellcolor[HTML]{E9AFAA} 2.712& 18.564\\
dec. time & 1.284 & 1.456 & 3.071 & \cellcolor[HTML]{B2B2FF} 0.554&\cellcolor[HTML]{E9AFAA}0.544\\
\bottomrule
\end{tabular}
\caption{Quantitative results on MVUB dataset. The best and the second best results are denoted by red and blue.}
\label{tab:comp_mvub}
\end{table}

\begin{figure}[htbp]
    \centering
    \includegraphics[width=\linewidth]{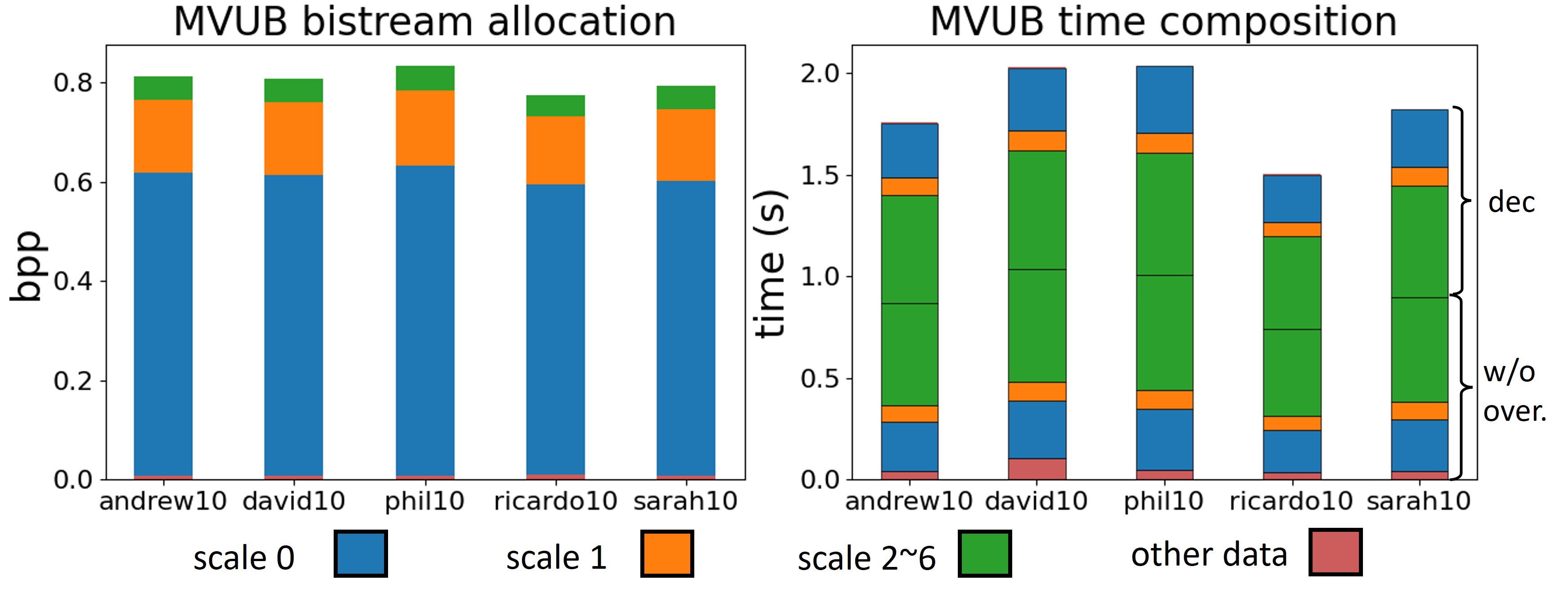}
    \caption{Bitstream allocation and time composition in MVUB. ``Other data" in the figure includes the size of network parameters together with the size of the lowest scale point cloud $pc_{low}$. And w/o over. denotes the encoding time without overfitting time.}
    \label{fig:compsition}
\end{figure}

\begin{table}[htp]
\centering
\small
\begin{tabular}{lccccc}
\toprule
\textbf{Metrics} & \textbf{Decoder} & $\boldsymbol{pc_{low}}$ & \textbf{scale 2-6} & \textbf{scale 1} & \textbf{scale 0} \\ \midrule
bpp & 0.73 & 0.17 & 5.83 & 18.10 & 75.17 
\\
enc. time & 0.47 & 8.58 & 30.47 & 14.92 & 45.56 
\\
dec. time & 0.52 & 0.00 & 31.60 & 16.25 & 51.63 
\\ \bottomrule
\end{tabular}
\caption{Statistics of bitstream proportion and encoding/decoding time proportion (\%) in MVUB.}
\label{tab:bpp_time_ratio}
\end{table}

\textbf{Bitstream allocation and time composition}.
We illustrate bitstream allocation and the time composition in \cref{fig:compsition}, and present the corresponding proportions in \cref{tab:bpp_time_ratio}. Bitstream allocation indicates that higher spatial scales result in larger bitstream consumptions, as point clouds at higher spatial scales contain more geometric details and thus require more information to predict occupancy. Network parameters occupy an ignorable proportion of the bitstream as they are shared across frames in a GoP. The time composition shows that most of the time is still spent in the hierarchical point cloud reconstruction process, rather than the encoding and decoding of network parameters and the lowest scale point cloud $pc_{low}$.


\subsection{Ablation Study} \label{sec:ablation}
We use the average performance on 8iVFB, Owlii and MVUB to evaluate the effectiveness of different settings of LINR-PCGC. The following figures and tables are shown on 8iVFB and MVUB. Tables and figures on Owlii dataset of this part are in the appendix.

\subsubsection{Ablation of initialization strategies} \label{sec:ab_is}
\begin{figure}[htbp]
    \centering
    \includegraphics[width=\linewidth]{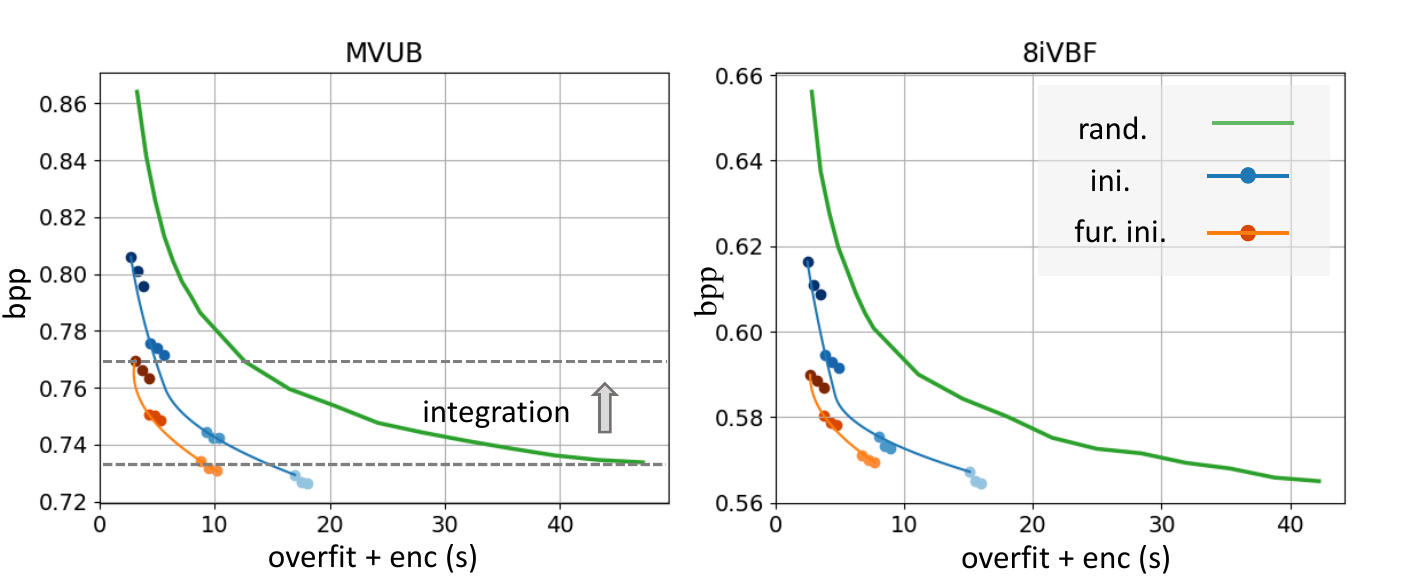}
    \caption{Training times vs. bpp curves with randomly initializing each GoP (rand.), randomly initialize the first GoP (ini.), and using similar sequences to initialize the first GoP (fur. ini.).}
    \label{fig:ab_pretrain}
\end{figure}

To demonstrate the acceleration effect of the initialization strategy on training, we set three different initialization methods: 1) random initialization for each GoP (rand.); 2) randomly initialize the first GoP and use the first GoP to initialize subsequent GoPs (ini.); and 3) use other similar sequences to initialize the first GoP, e.g., \emph{basketball} initializes the first GoP of \emph{dancer}, and use the first GoP to initialize subsequent GoPs (fur. ini.). \cref{fig:ab_pretrain} shows training time vs. bpp curves of the three initialization methods. And by integrating the overlapping parts of the three curves along the bpp axis, we can estimate the average training time for the three methods and calculate the time ratios shown in \cref{tab:ab_pretrain}. Leveraging the correlation between GOPs and similarity among sequences can significantly improve training efficiency, i.e., 65.3\% and 76.0\% of average time saving for method 2) and 3) compared to method 1). 

\begin{table}[htbp]
\centering
\small
\begin{tabular}{lccccc}
\toprule
 & \textbf{8iVFB} & \textbf{Owlii} & \textbf{MVUB} & \textbf{avg.}  \\
\midrule
\textbf{ini.} & 36.0 & 34.4 & 33.7 & 34.7  \\
\textbf{fur. ini.} & 22.9 & 29.2 & 20.0& 24.0 \\
\bottomrule
\end{tabular}
\caption{Relative time (\%) that ini. and fur. ini. take compared to rand.}
\label{tab:ab_pretrain}
\end{table}

\subsubsection{Ablation of modules} 
\begin{figure}[ht]
    \centering
    \includegraphics[width=\linewidth]{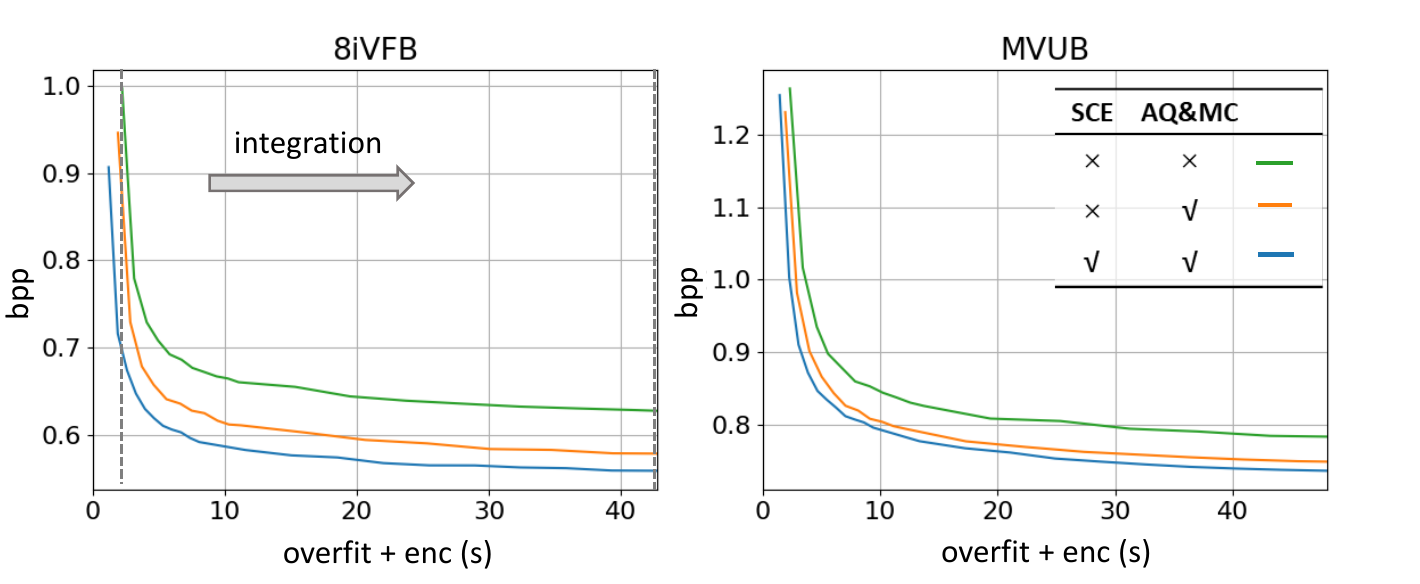}
    \caption{Impact of each module in LINR-PCGC.} 
    \label{fig:ab_module}
\end{figure}

To demonstrate the effectiveness of each module, we start by retaining only the CNP module and then sequentially adding other modules until the complete LINR-PCGC without the initialization strategy is implemented. The results are shown in \cref{fig:ab_module}. Then we integrate the overlapping parts of all the curves over time to obtain the average bpp ratios relative to the green curve, as shown in \cref{tab:ab_module}. AQ and MC modules can reduce 8.1\% bpp, while SCE can further reduce 3.1\% bpp, indicating the effectiveness of the proposed modules.

\begin{table}[htbp]
\centering
\begin{tabular}{cccc}
\toprule
SCE & AQ\&MC & r.t. bpp (\%)$\downarrow$ \\ \midrule
\texttimes & \texttimes &  100.0 \\ 
\texttimes & \checkmark &  91.9 \\ 
\checkmark & \checkmark &  88.8 \\ 

\bottomrule
\end{tabular}
\caption{Impact of each module in LINR-PCGC, where r.t. bpp denotes relative bpp (\%) over the method w/o. SCE, AQ\&MC.}
\label{tab:ab_module}
\end{table}

\subsection{Conclusion}
We propose LINR-PCGC to compress a sequence of point clouds with the basic architecture of INR methods. So, our method inherits the biggest advantage of INR: it does not rely on specific data distributions to work. Additionally, we employ an initialization strategy for acceleration and thus achieve an encoding time comparable to non-INR methods. In addition, the lightweight network design ensures a shorter decoding time. 
Further, we will include inter-frame prediction for temporal redundancy removal and extend to lossy compression as our method reduces restrictions of network size in INR methods.

{
    \small
    \bibliographystyle{ieeenat_fullname}
    \bibliography{main}
}
\clearpage
\setcounter{page}{1}
\maketitlesupplementary
\setcounter{section}{0}
\section{Appendix}

\subsection{Detail of parameters}
The details of the parameters in our experiment are listed in \cref{tab:parameters}.

\begin{table}[h]
\centering
\begin{tabularx}{\linewidth}{|c|>{\hsize=0.6\hsize}X|c|}
\hline
\textbf{Symbol} & \textbf{Description} & \textbf{Value} \\ \hline
$lr_0$ & Initial learning rate & 0.01 \\ \hline
$lr_{min}$ & Minimum learning rate & 0.0004 \\ \hline
$\gamma$ & Multiplicative factor of learning rate decay in StepLR & 0.992 \\ \hline
$\text{step size}$ & Period of learning rate decay in StepLR & 32 \\ \hline
$\lambda$ & Weight decay (for L2 penalty) factor in Adam & 0.0001 \\ \hline
$epoch_f$ & Training epoch number for the first GoP & 6-60 \\ \hline
$epoch_s$ & Training epoch number for subsequent GoPs & 6-60 \\ \hline
$T$ & GoP size & 32 \\ \hline
$M$ & Total frame count of an entire testing sequence & 96 \\ \hline
$C_{mlp}$ & Hidden channel dimension of the MLP & 24 \\ \hline
$C_{sconv}$ & Hidden channel dimension of the SConv & 8 \\ \hline
$C_{EMB}$ & Channel dimension of the SEMB & 8 \\ \hline
\end{tabularx}
\caption{Detail of parameters of our experiment.}
\label{tab:parameters}
\end{table}

\begin{figure*}
    \centering
    \includegraphics[width=\linewidth]{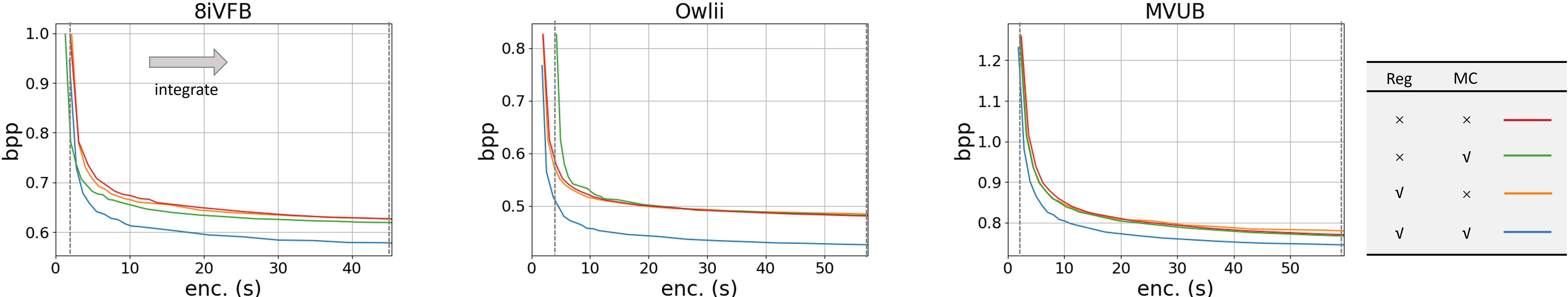}
    \caption{The impact of regularization terms on MC modules.}
    \label{fig:reg_impact}
\end{figure*}

\subsection{Supplementary Experiment result}
To provide a quantitative analysis on the number of training epochs for the first GoP $F$, detailed bpp values under different $F$ are given in \cref{tab:8i_append,tab:owlii_append,tab:mvub_append}, where the training epoch for subsequent GoPs is fixed to 1, bpp denotes the average bpp of all sequences in a dataset, r.t. bpp denotes the relative bpp (\%) of other methods over G-PCC, and w/o over. denotes the encoding time without overfitting time of our method. All times are in seconds.

\begin{table}[htbp]
\centering
\begin{tabular}{lcccc}
\toprule
 & F:6 & F:11 & F:31 & F:61 \\
\midrule
longdress & 0.618 & 0.597 & 0.576 & 0.573 \\
loot & 0.57 & 0.549 & 0.532 & 0.528 \\
redandblack & 0.689 & 0.665 & 0.64 & 0.629 \\
soldier & 0.588 & 0.567 & 0.553 & 0.539 \\
\midrule
bpp & 0.616 & 0.594 & 0.576 & 0.567 \\
r.t. bpp & 82.925 & 79.975 & 77.422 & 76.311 \\
w/o over.  & 0.477 & 0.512 & 0.446 & 0.44 \\
enc. time  & 2.464 & 3.869 & 8.005 & 15.092 \\
dec. time  & 0.501 & 0.535 & 0.471 & 0.465 \\
\bottomrule
\end{tabular}
\caption{Quantitative results on 8iVFB dataset of different $F$.}
\label{tab:8i_append}
\end{table}

\begin{table}[htbp]
\centering
\begin{tabular}{lcccc}
\hline
 & F:6 & F:11 & F:31 & F:61 \\
\hline
basketball & 0.452 & 0.438 & 0.422 & 0.414 \\
dancer & 0.473 & 0.457 & 0.44 & 0.432 \\
exercise & 0.46 & 0.443 & 0.428 & 0.418 \\
model & 0.475 & 0.461 & 0.445 & 0.434 \\
\midrule
bpp & 0.465 & 0.45 & 0.434 & 0.425 \\
r.t. bpp & 78.759 & 76.204 & 73.497 & 71.949 \\
\midrule
\hline
w/o over. & 0.402 & 0.46 & 0.389 & 0.397 \\
enc. time & 2.071 & 3.392 & 6.972 & 12.958 \\
dec. time & 0.422 & 0.478 & 0.41 & 0.417 \\
\hline
\end{tabular}
\caption{Quantitative results on Owlii dataset of different $F$.}
\label{tab:owlii_append}
\end{table}

\begin{table}[htbp]
\centering
\begin{tabular}{lcccc}
\hline
 & F:6 & F:11 & F:31 & F:61 \\
\hline
andrew10 & 0.833 & 0.801 & 0.769 & 0.754 \\
david10 & 0.778 & 0.758 & 0.723 & 0.708 \\
phil110 & 0.841 & 0.812 & 0.777 & 0.772 \\
ricardo10 & 0.802 & 0.76 & 0.729 & 0.709 \\
sarah10 & 0.777 & 0.748 & 0.724 & 0.704 \\
\midrule
bpp & 0.806 & 0.776 & 0.744 & 0.729 \\
r.t. bpp & 87.548 & 84.25 & 80.844 & 79.206 \\
w/o over & 0.524 & 0.57 & 0.524 & 0.518 \\
enc. time & 2.712 & 4.385 & 9.291 & 16.967 \\
dec. time & 0.554 & 0.599 & 0.554 & 0.548 \\
\hline
\end{tabular}
\caption{Quantitative results on MVUB dataset of different $F$.}
\label{tab:mvub_append}
\end{table}

\subsection{Supplementary Bitstream and Time Allocation}
\begin{figure}[htbp]
    \centering
    \includegraphics[width=\linewidth]{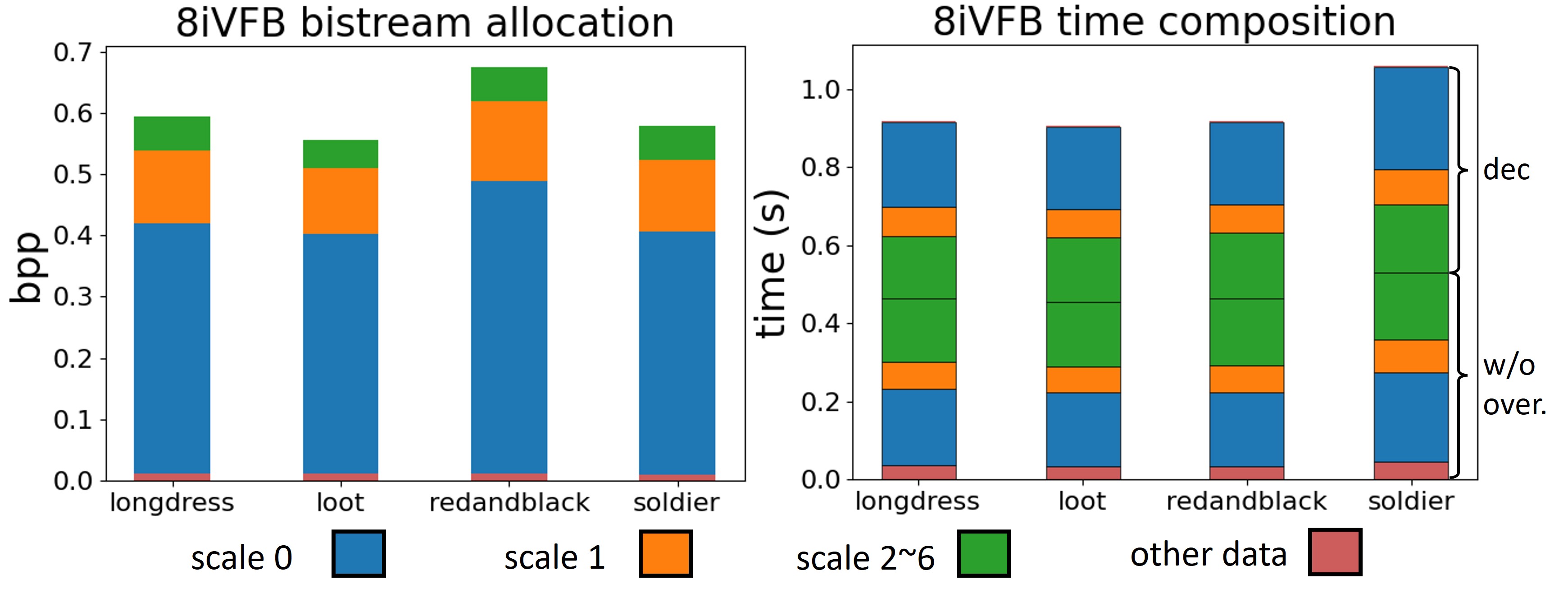}
    \caption{Bitstream allocation and encoding/decoding time composition in 8iVFB.}
    \label{fig:compsition_8i}
\end{figure}

\begin{figure}[htbp]
    \centering
    \includegraphics[width=\linewidth]{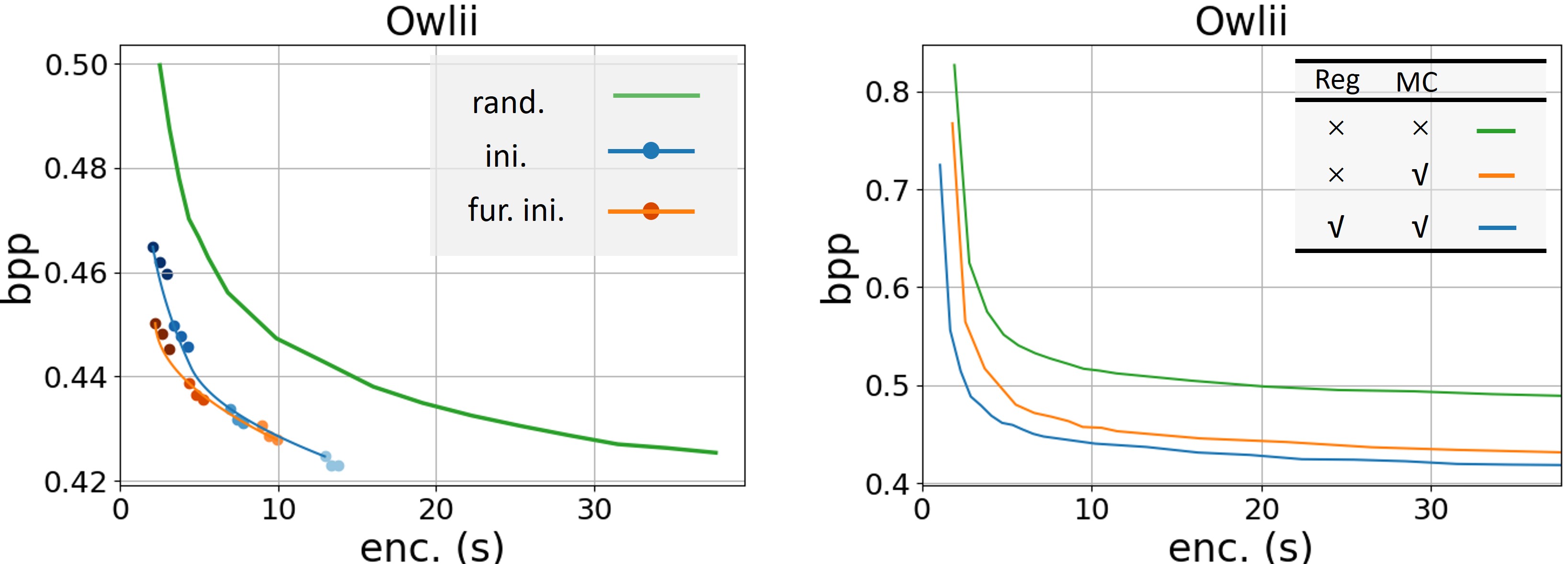}
    \caption{(a) The training time--bpp curves with randomly initializing each GoP (rand.), ini., and fur. ini. in Owlii. (b) Impact of each module in LINR-PCGC in Owlii.}
    \label{fig:ablation_owlii}
\end{figure}

In \cref{sec:exp_result}, we have given the bitstream allocation and the encoding/decoding time composition figures of MVUB. Here we give the figures for 8iVFB and Owlii as \cref{fig:compsition_8i,fig:ablation_owlii}. \cref{sec:ablation} has given the training times vs. bpp curves of 8IVFB and MVUB, and here we give the curves of Owlii in \cref{fig:ablation_owlii}. 

\begin{figure}[htbp]
    \centering
    \includegraphics[width=\linewidth]{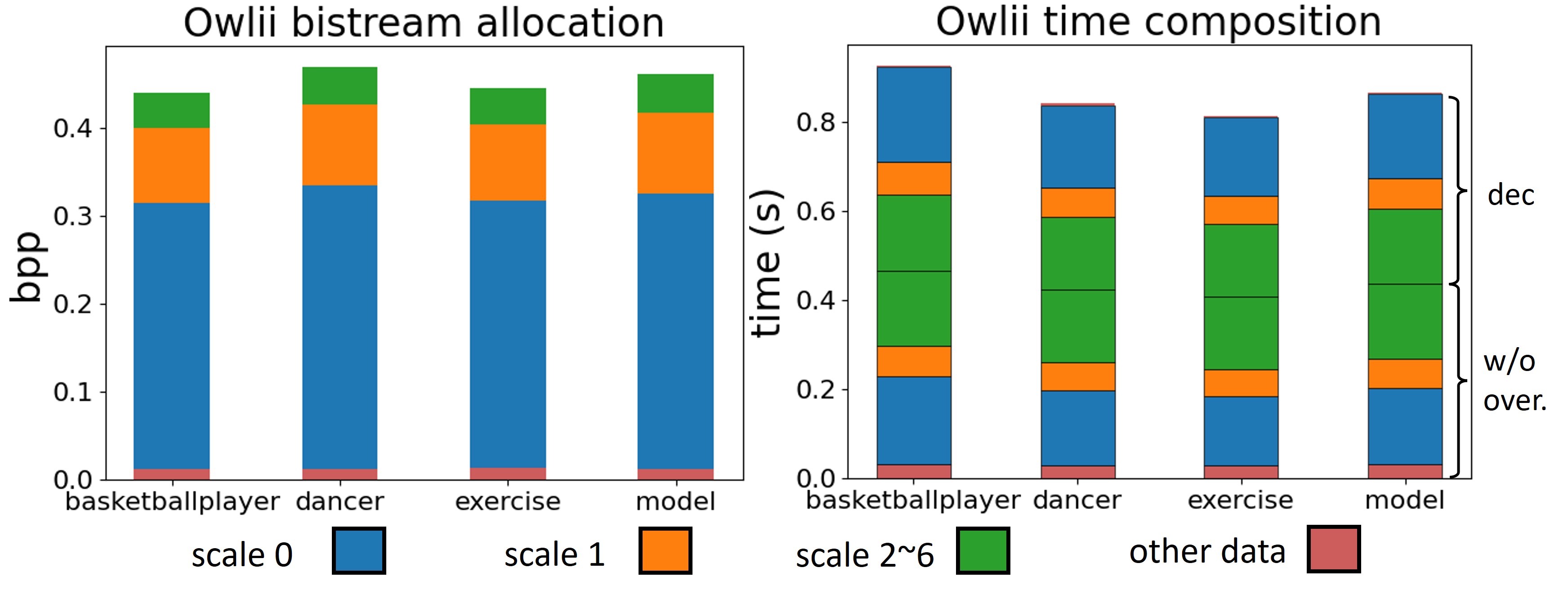}
    \caption{Bitstream allocation and encoding/decoding time composition in Owlii.}
    \label{fig:compsition_owlii}
\end{figure}

\subsection{Supplementary of Ablation Study}
\begin{figure}
    \centering
    \includegraphics[width=\linewidth]{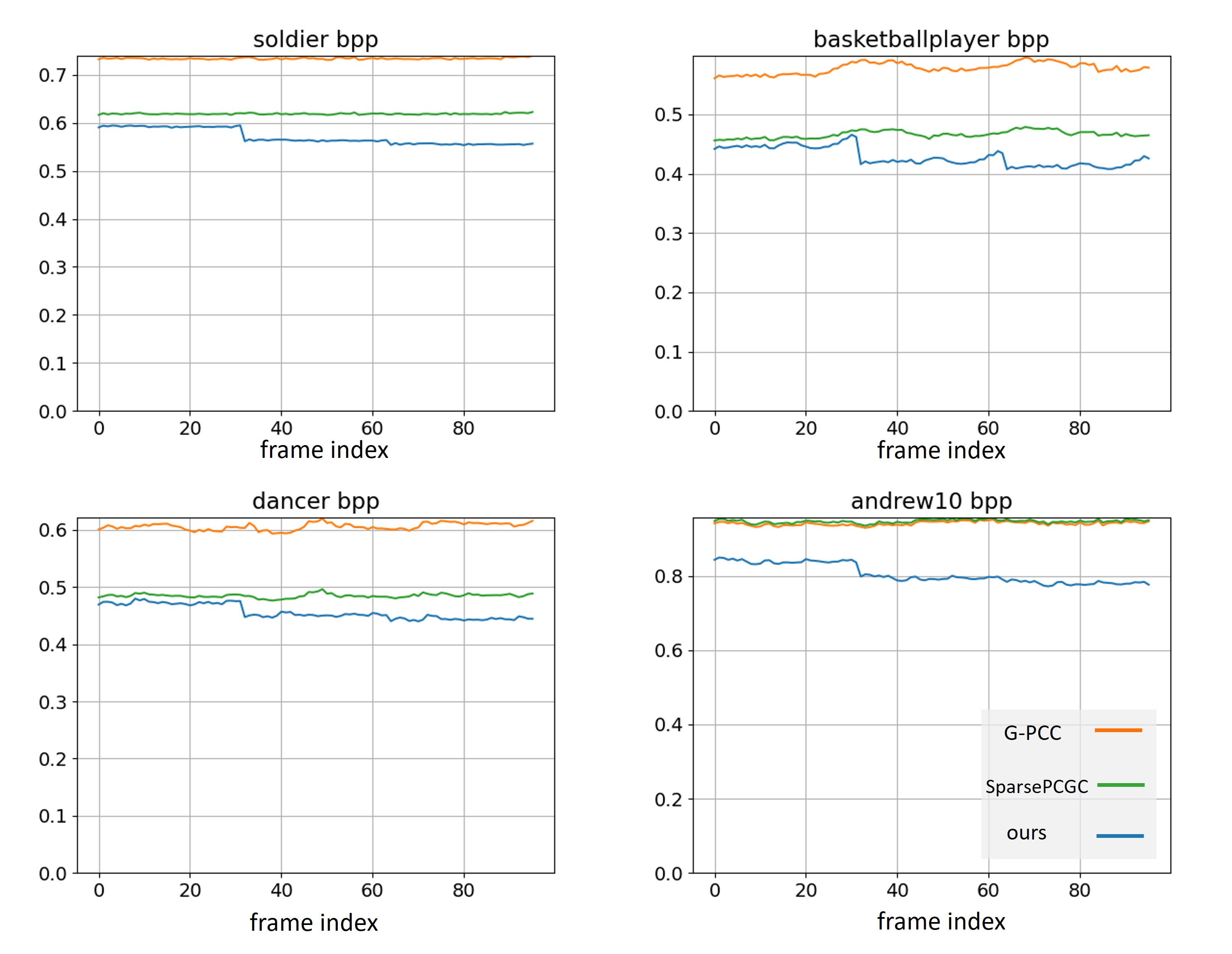}
    \caption{Bitstream size of each frame. The frame number of each sequence is 96, and GoP size is 32. Both the first GoP and subsequent GoPs are trained for 6 epochs.}
    \label{fig:frame_serials}
\end{figure}

\textbf{Supplementary analysis of initialization strategy.} To further demonstrate the effectiveness of the initialization strategy, we have sketched \cref{fig:frame_serials}. We can observe from the figure that the bitstream of each GoP has a significant decrease. This is because the parameters of the latter GoP are initialized by the parameters of the previous GoP. Under the same optimization time, the later the GoP, the better the encoding efficiency can be achieved. 

\textbf{The effect of Model Compression (MC).} To demonstrate the effectiveness of MC, we presented the original size of the network parameters (Ori.), the bitstream size of directly converting quantized integer parameters into a bitstream (ToByte), the bitstream size after further compressing ToByte using LZ77 (LZ77), and the bitstream size generated by arithmetic coding using a Laplace distribution (Laplace). As depicted in \cref{tab:param_sizes}, arithmetic coding with a Laplacian prior assumption significantly reduces the bitstream size of model parameters.
\begin{table}[htbp]
\centering
\begin{tabular}{lcccc}
\toprule
 & \textbf{Owlii} & \textbf{8iVFB} & \textbf{MVUB} & \textbf{Avg}\\
\midrule
\textbf{Ori.} & 1750784 & 1750784 & 1750784 &1750784 
\\
\textbf{ToByte} & 437762 & 437762 & 437762 &437762 
\\
\textbf{LZ77} & 267790 & 269554 & 250825.4 &268672 
\\
\textbf{Laplace} & 248490 & 251360 & 240352.9 &251360 
\\
\bottomrule
\end{tabular}
\caption{Bitstream sizes (in bits) of the model parameters under different model compression algorithms.}
\label{tab:param_sizes}
\end{table}

\textbf{The effect of the regularization item (Reg.).} The regularization term can reduce the absolute value of the network parameters, thus making the quantized parameters closer to the Laplace distribution. Therefore, adding a regularization term is beneficial for MC. The specific situation is shown in \cref{fig:reg_impact}. We choose the method with Reg. and MC as the baseline. Then, we integrate the overlapping parts of time and make a ratio to the baseline to obtain \cref{tab:reg_impact}. We can observe from the first and second lines that when there is no regularization term, the MC module can only save 0.826\% of the bitstream. Next, we can observe from the third and fourth lines of the table that when there exists a regularization term, MC can save 8.17\% bitstream. Although we can conclude from the comparison between the first and third lines that the presence of regularization terms alone does not result in significant stream savings (0.092\%), its existence is one of the foundations for the functioning of the MC module.

\begin{figure}
    \centering
    \includegraphics[width=\linewidth]{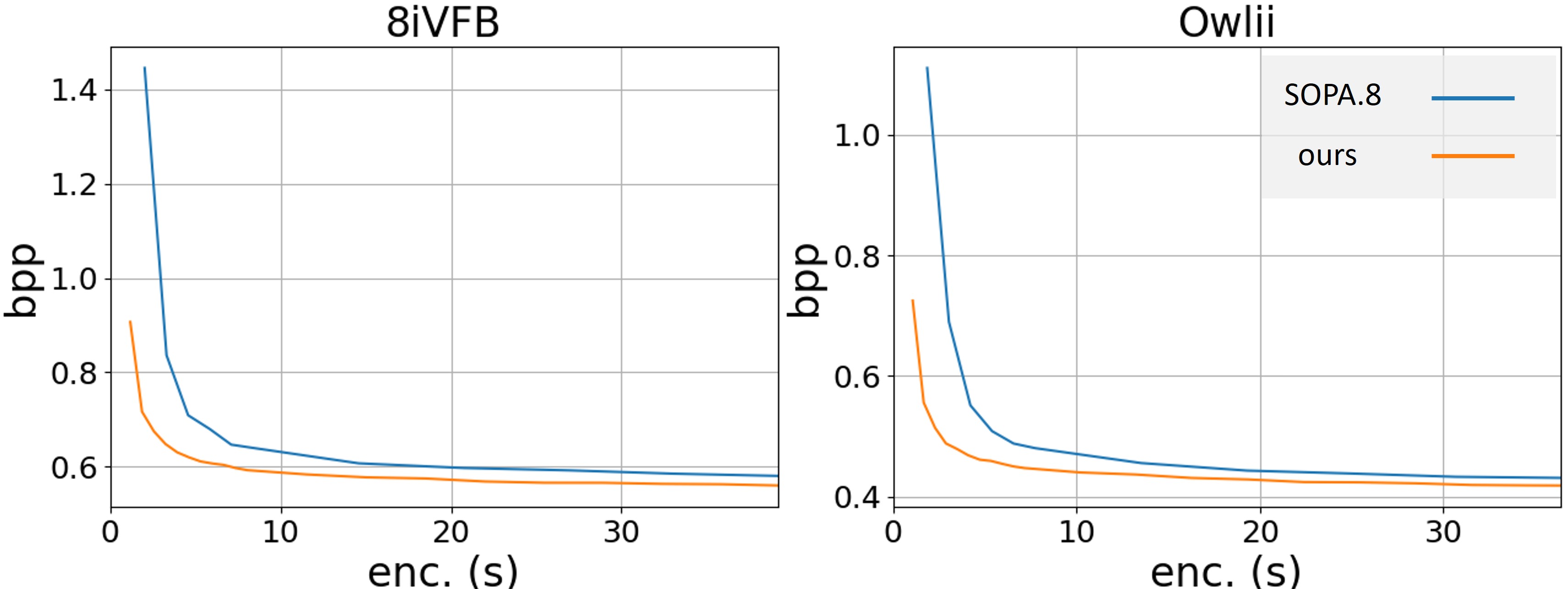}
    \caption{Comparison of ours method (without pretrain) and replace our CNP module to 8-stage SOPA with 8 hidden channels.}
    \label{fig:cmp_sopa}
\end{figure}

\textbf{The advantages of CNP under the INR architecture.} The simplest idea for upsampling is to directly use SOPA from SparsePCGC and perform overfitting. However, SOPA training takes nearly 9 hours and has tens of millions of bits of parameters\footnote{This information comes from the training log provided by the authors.}. For online training, this is expensive and unacceptable. Therefore, we reduce the number of hidden channels in the 8-stage SOPA from 32 to 8 and utilize channel-wise prediction to replace transpose SparseConv. And we illustrate the comparison result in \cref{fig:cmp_sopa}. Then we integrate the overlapping parts of time and calculate the ratio relative to SOPA to obtain \cref{tab:cmp_sopa}. From the table, we can observe that CNP can save about 7.62\% of the bitstream compared to 8-stage SOPA. To further demonstrate the advantages of CNP under the INR architecture, we construct \cref{tab:memory_cmp} which shows the comparison between CNP and 8-stage SOPA. From the table, we can observe that CNP can save approximately 61.91\% of peak memory with the same number of hidden channels. This also indicates that prediction based on a two-layer octree structure is more memory efficient than transpose convolution in SOPA.\footnote{We did not compare on MVUB because running 8-stage SOPA with 8 hidden channels on the MVUB dataset would exceed the memory of RTX 3090 in our INR framework.}

\begin{table}[h]
\centering
\begin{tabular}{lccc}
\toprule
 & Owlii & 8iVFB & avg \\
\midrule
8-stage SOPA & 1 & 1 & 1 \\
ours & 0.9238 & 0.9238 & 0.9238 \\
\bottomrule
\end{tabular}
\caption{Comparison of ours method (without pretrain) and replace our CNP module to 8-stage SOPA with 8 hidden channels.}
\label{tab:cmp_sopa}
\end{table}

\begin{table}[h]
\centering
\begin{tabular}{lccc}
\toprule
 & Owlii & 8iVFB & avg \\
\midrule
8-stage SOPA & 10.64 & 13.21 & 11.92 \\
ours & 4.00 & 5.09 & 4.54 \\
\bottomrule
\end{tabular}
\caption{Comparison of peak memory usage between ours method and 8-stage SOPA with 8 hidden channels.}
\label{tab:memory_cmp}
\end{table}

\begin{table}[h]
\centering
\begin{tabular}{cccccc}
\toprule
Reg & MC & Owlii & 8iVFB & MVUB & avg. \\
\midrule
$\times$ & $\times$ & 1.132 & 1.089 & 1.049 & 1.090 \\
$\times$ & \checkmark & 1.142 & 1.062 & 1.041 & 1.081 \\
\checkmark & $\times$ & 1.132 & 1.085 & 1.050 & 1.089 \\
\checkmark & \checkmark & 1.000 & 1.000 & 1.000 & 1.000 \\
\bottomrule
\end{tabular}
\caption{The impact of regularization terms on MC modules.}
\label{tab:reg_impact}
\end{table}

\subsection{Bitstream heatmap}
\begin{figure}
    \centering
    \includegraphics[width=\linewidth]{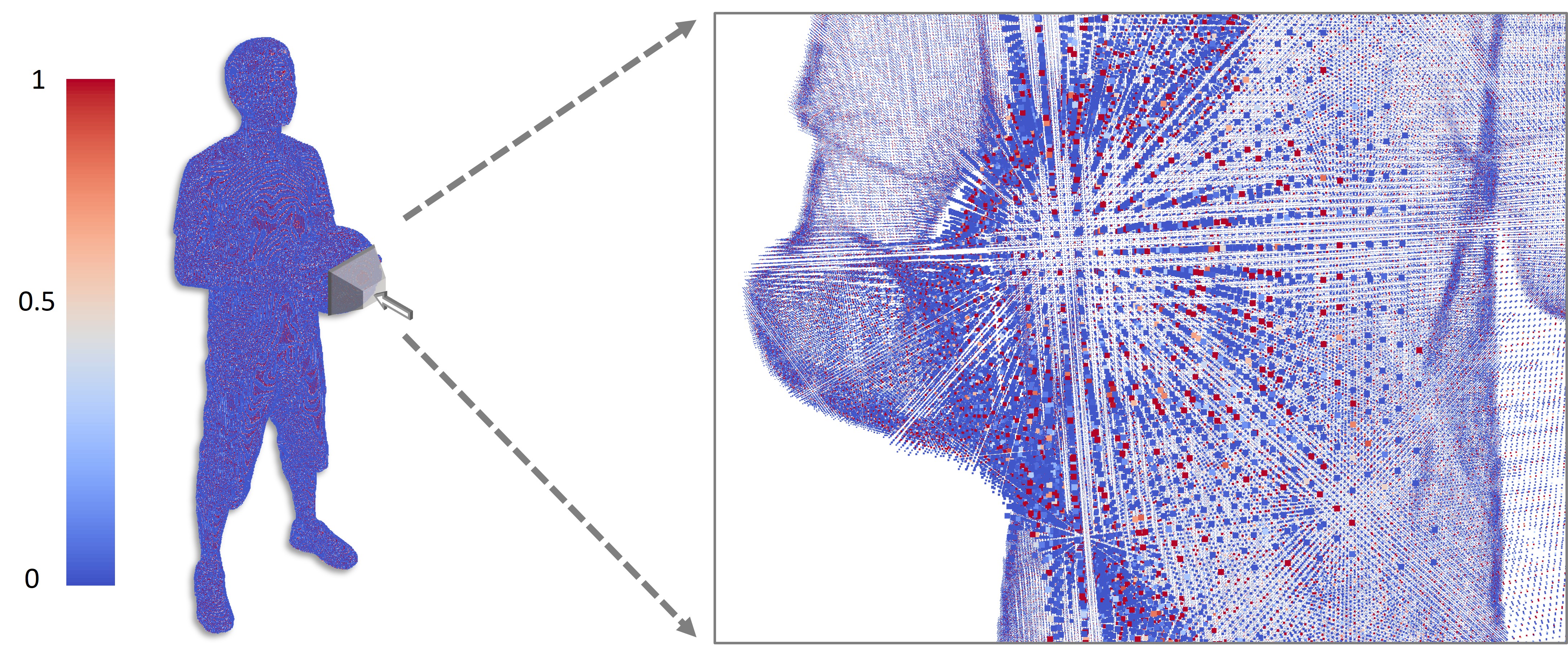}
    \caption{Bitstream heatmap. }
    \label{fig:pc_show}
\end{figure}

\cref{fig:pc_show} illustrates the absolute difference between the estimated occupancy probability and the actual occupancy values. A larger difference is equivalent to a higher bitrate of a point. Points with higher bit rates appear periodically in the size of $2^3$ cubes as the right part of \cref{fig:pc_show}. This phenomenon occurs because we use decoded child nodes to predict non-decoded child nodes. The first batch of child nodes typically has higher bit rates due to the lack of current scale priors, while those predicted based on other child nodes have lower bit rates.

\end{document}